\documentclass{article} 
\usepackage{nips14submit_e,times}
\usepackage{hyperref}
\usepackage{url}
\usepackage{sidecap}

\usepackage{times}
\usepackage[pdftex]{graphicx}
\usepackage{subfigure} 

\usepackage{dsfont}
\usepackage{amsmath}

\usepackage{wasysym}

\usepackage{etoolbox}

\usepackage[numbers,square]{natbib}

\patchcmd{\footnotemark}{\stepcounter{footnote}}{\refstepcounter{footnote}}{}{}
\newcommand{\ignore}[1]{}

\newenvironment{itemizesquish}{\begin{list}{\labelitemi}{\setlength{\itemsep}{0em}\setlength{\labelwidth}{0.5em}\setlength{\leftmargin}{\labelwidth}\addtolength{\leftmargin}{\labelsep}}}{\end{list}}
\newcommand{\obs}{x}
\newcommand{\obsdomain}{\cal{X}}
\newcommand{\rec}{\hat{x}}
\newcommand{\recdomain}{\hat{{\cal{X}}}}
\newcommand{\hid}{y}
\newcommand{\hiddomain}{\cal{Y}}
\newcommand{\sid}{\phi}

\newcommand{\thin}[1]{\textsf{\small{#1}}}

\newcommand{\hmm}{\thin{hmm}}
\newcommand{\fhmm}{\thin{fhmm}}
\newcommand{\auto}{\thin{auto}}
\newcommand{\giza}{\thin{mgiza++}}
\newcommand{\fastalign}{\thin{fast\_align}}
\newcommand{\modelfour}{\thin{model 4}}
\newcommand{\BigO}[1]{\ensuremath{\operatorname{O}\left(#1\right)}}

\title{Conditional Random Field Autoencoders \\ for Unsupervised Structured Prediction}

\author{Chu-Cheng Lin \qquad Waleed Ammar \qquad Lori Levin \qquad Chris Dyer \\
        Language Technologies Institute \\
        Carnegie Mellon University \\
        Pittsburgh, PA, 15213, USA \\
        {\tt \{chuchenl,wammar,lsl,cdyer\}@cs.cmu.edu} }

\author{
Waleed Ammar \qquad Chris Dyer \qquad Noah A.~Smith \\
School of Computer Science \\
Carnegie Mellon University \\
Pittsburgh, PA 15213, USA \\
{\tt \{wammar,cdyer,nasmith\}@cs.cmu.edu}
}

\nipsfinalcopy 

\begin{document}

\maketitle

\begin{abstract}
We introduce a framework for unsupervised learning of structured predictors with overlapping, global features. 
Each input's latent representation is predicted conditional on the observed data using a feature-rich conditional random field (CRF). 
Then a reconstruction of the input is (re)generated, conditional on the latent structure, using a generative model which factorizes similarly to the CRF.
The autoencoder formulation enables efficient exact inference without resorting to unrealistic independence assumptions or restricting the kinds of features that can be used.
We illustrate connections to traditional autoencoders, posterior regularization, and multi-view learning.
We then show competitive results with instantiations of the framework for two canonical tasks in natural language processing: part-of-speech induction and bitext word alignment, and show that training the proposed model can be substantially more efficient than a comparable feature-rich baseline.
\end{abstract}

\section{Introduction}
Conditional random fields~\cite{lafferty:01} are used to model structure in numerous problem domains, including natural language processing (NLP), computational biology, and  computer vision. They enable efficient inference while incorporating rich features that capture useful domain-specific insights. Despite their ubiquity in supervised settings, CRFs---and, crucially, the insights about effective feature sets obtained by developing them---play less of a role in \emph{unsupervised} structure learning, a problem which traditionally requires jointly modeling observations and the latent structures of interest. For unsupervised structured prediction problems, less powerful models with stronger independence assumptions are standard.\footnote{For example, a first-order hidden Markov model requires that $y_i \perp x_{i+1} \mid y_{i+1}$ for a latent sequence $\mathbf{y} = \langle y_1, y_2, \ldots \rangle$ generating $\mathbf{x} = \langle x_1, x_2, \ldots\rangle$, while a first-order CRF allows $y_i$ to directly depend on $x_{i+1}$.}   This state of affairs is suboptimal in at least three ways: (i) adhering to inconvenient independence assumptions when designing features is limiting---we contend that effective feature engineering is a crucial mechanism for incorporating inductive bias in unsupervised learning problems; (ii) features and their weights have different semantics in joint and conditional models (see \S\ref{sec:alternatives}); and (iii) modeling the generation of high-dimensional observable data with feature-rich models is computationally challenging, requiring expensive marginal inference in the inner loop of iterative parameter estimation algorithms (see \S\ref{sec:alternatives}).

Our approach leverages the power and flexibility of CRFs in unsupervised learning without sacrificing their attractive computational properties or changing the semantics of well-understood feature sets. Our approach replaces the standard joint model of observed data and latent structure with a two-layer \textbf{conditional random field autoencoder} that first generates latent structure with a CRF (conditional on the observed data) and then (re)generates the observations conditional on just the predicted structure. For the reconstruction model, we use distributions which offer closed-form maximum likelihood estimates  (\S\ref{sec:model}). The proposed architecture provides several mechanisms for encouraging the learner to use its latent variables to find intended (rather than common but irrelevant) correlations in the data. First, hand-crafted feature representations---engineered using knowledge about the problem---provide a key mechanism for incorporating inductive bias. Second, by reconstructing a transformation of the structured input. Third, it is easy to simultaneously learn from labeled and unlabeled examples in this architecture, as we did in \cite{lin:14}. In addition to the modeling flexibility, our approach is computationally efficient in a way achieved by no other unsupervised, feature-based model to date:  under a set of mild independence assumptions regarding the reconstruction model, per-example inference required for learning in a CRF autoencoder is no more expensive than in a supervised CRF with the same independence assumptions.

In the next section we describe the modeling framework, then review related work and show that, although our model admits more powerful features, the inference required for learning is simpler (\S\ref{sec:related}). We conclude with experiments showing that the proposed model achieves state-of-the-art performance on two  unsupervised tasks in NLP: POS induction and word alignment, and find that it is substantially more efficient than MRFs using the same feature set (\S\ref{sec:exp}).

\begin{figure}
\begin{minipage}{.46\textwidth}
  \includegraphics[height=1.2in]{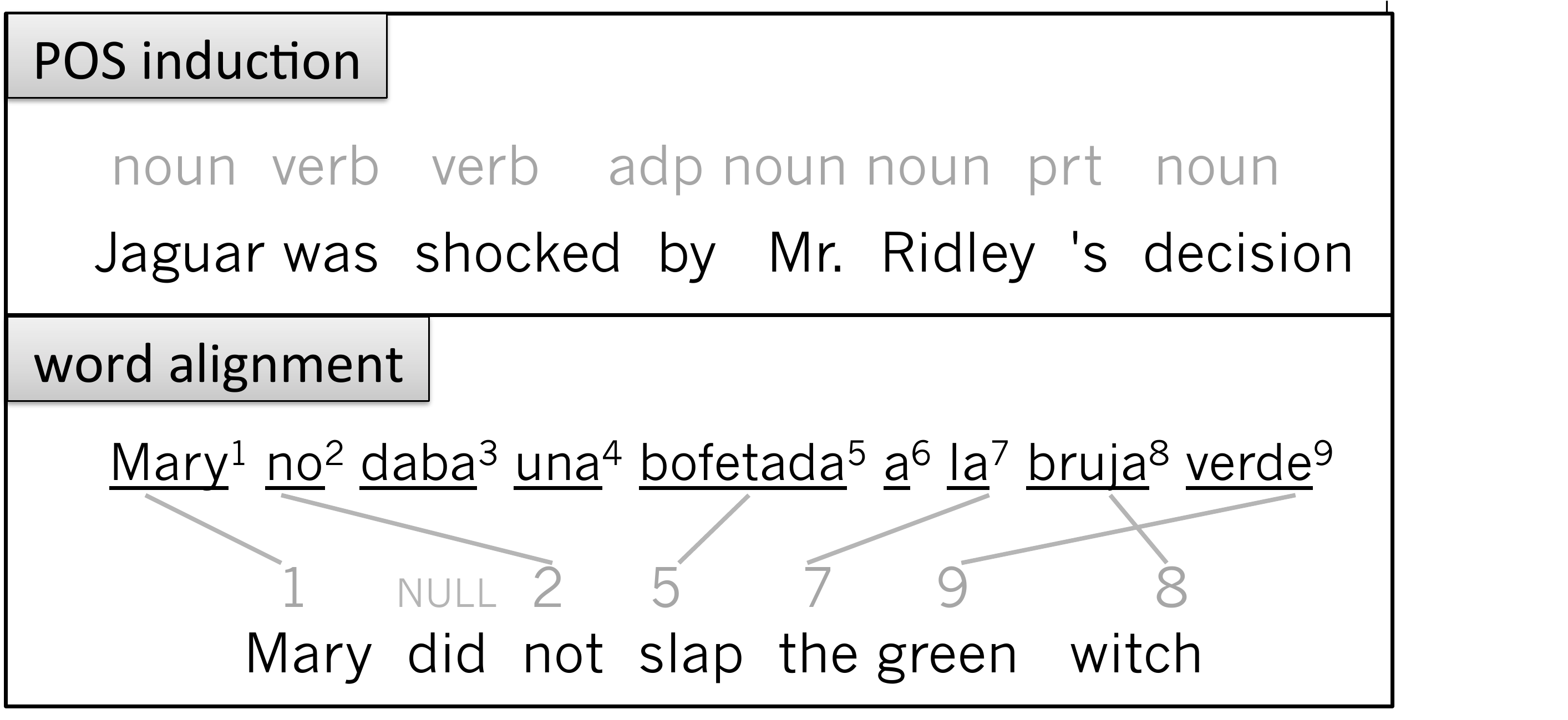}
\end{minipage}%
\begin{minipage}{.5\textwidth}
\noindent  \begin{tabular}{cll}
    & \textbf{POS induction} & \textbf{Word alignment} \\ 
    $\cal{T}$    &  corpus of sentences & parallel corpus \\ 
    $\mathbf{\obs}$    & sentence & target sentence \\ 
    $\obsdomain$    & vocabulary & target vocabulary \\ 
    $\boldsymbol{\sid}$    & (opt.) tag dictionary & source sentence \\ 
    $\mathbf{\hid}$    & POS sequence & sequence of indices \\
    &                  &in the source sentence \\ 
    $\hiddomain$    & POS tag set & $\{\textsc{null},1, 2, \ldots, |\boldsymbol{\phi}|\}$ \\
  \end{tabular}
\end{minipage}
\caption{Left: Examples of structured observations (in black), hidden structures (in {\color[rgb]{0.5,0.5,0.5} gray}), and side information (\underline{underlined}). Right: Model variables for POS induction (\S\ref{sec:pos}) and word alignment (\S\ref{sec:wa}).  A parallel corpus consists of pairs of sentences (``source'' and ``target''). \label{fig:examples} \label{table:spec}}
\end{figure}

\section{Conditional Random Field Autoencoder}
\label{sec:model}
Given a training set ${\cal{T}}$ of observations (e.g., sentences or pairs of sentences that are translationally equivalent), consider the problem of inducing the hidden structure in each observation.  Examples of hidden structures include shallow syntactic properties (part-of-speech or POS tags), correspondences between words in translation (word alignment), syntactic parses and morphological analyses.

\paragraph{Notation.} Let each observation be denoted $\mathbf{\obs} = \langle \obs_1, \ldots, \obs_{|\mathbf{\obs}|}\rangle \in \obsdomain^{|\mathbf{\obs}|}$, a variable-length tuple 
  of discrete variables, $\obs \in \obsdomain$.
The hidden variables $\mathbf{\hid} = \langle \hid_1, \ldots, \hid_{|\mathbf{\hid}|} \rangle \in \hiddomain^{|\mathbf{\hid}|}$ form a tuple whose length is determined by $\mathbf{\obs}$, also taking discrete values.\footnote{In the interest of notational simplicity, we conflate random variables with their values.}  
We assume that $|\hiddomain|^{|\mathbf{\hid}|} \ll |\obsdomain|^{|\mathbf{\obs}|}$, which is typical in structured prediction problems. 
Fig.~\ref{table:spec} (right) instantiates $\mathbf{\obs}$, $\obsdomain$, $\mathbf{\hid}$, and $\hiddomain$ for two NLP tasks. 

\begin{figure}
\centering
\begin{minipage}{.5\textwidth}
  \centering
  \includegraphics[height=1.7in]{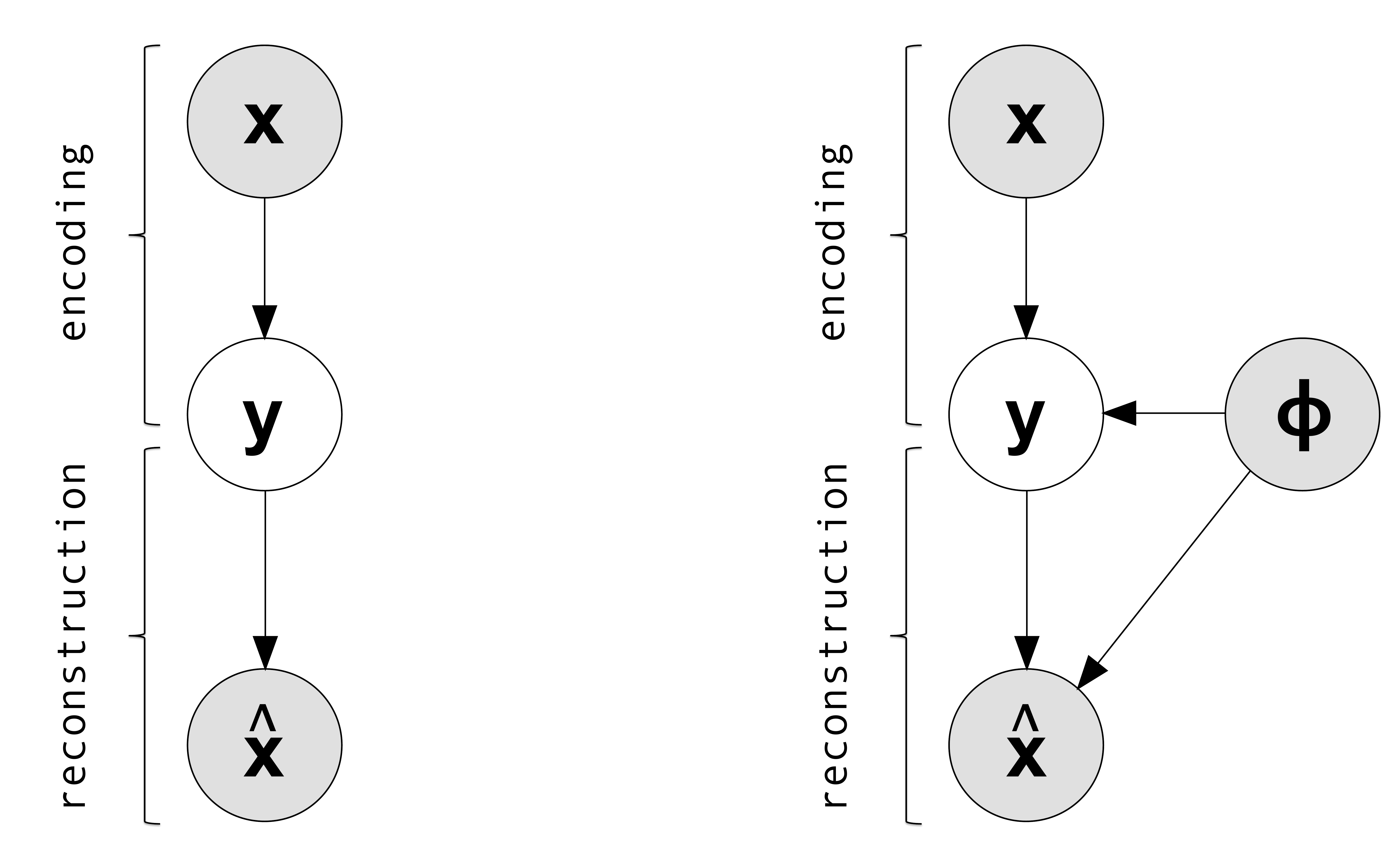}
\end{minipage}%
\begin{minipage}{.5\textwidth}
  \centering
  \includegraphics[height=1.7in]{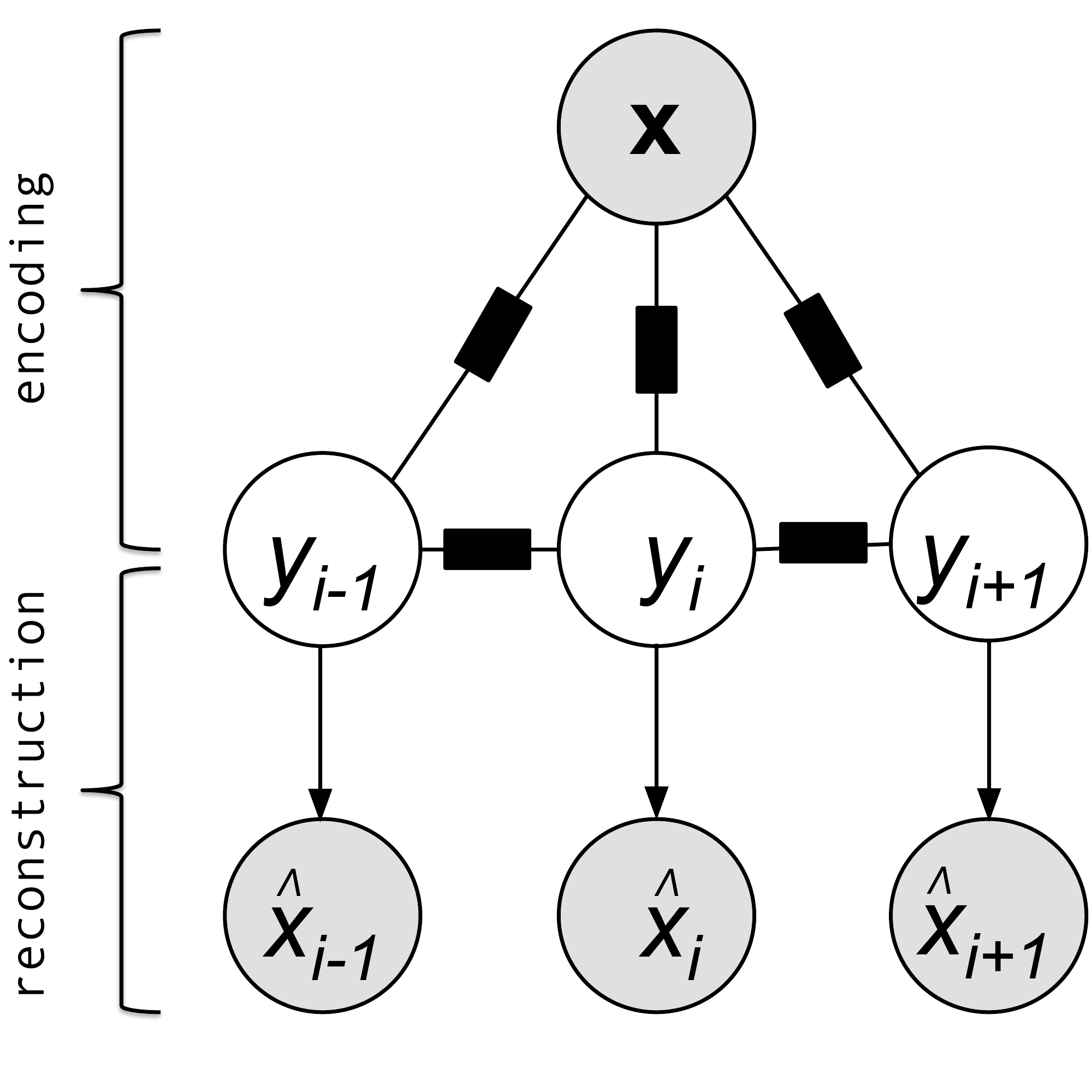}
\end{minipage}
\caption{ Graphical model representations of CRF autoencoders. 
  Left: the basic autoencoder model where the observation $\mathbf{\obs}$ generates the hidden structure $\mathbf{\hid}$ (encoding), which then generates $\mathbf{\rec}$ (reconstruction).
  Center: side information ($\sid$) is added.
  Right: a factor graph showing first-order Markov dependencies among elements of the hidden structure $\mathbf{\hid}$. \label{fig:model}  }
\end{figure}

Our model introduces a new observed variable, $\mathbf{\rec} = \langle \rec_1, \ldots, \rec_{|\mathbf{\rec}|}\rangle$.  
In the basic model in Fig.~\ref{fig:model} (left), $\mathbf{\rec}$ is a copy of $\mathbf{\obs}$.
The intuition behind this model is that a good hidden structure should be a likely \emph{encoding} of observed data and should permit \emph{reconstruction} of the data with high probability.

\paragraph{Sequential latent structure.} 
In this paper, we focus on \emph{sequential} latent structures with first-order Markov properties, i.e., $\hid_i \perp \hid_j \mid \{\hid_{i-1}, \hid_{i+1}\}$, as illustrated in Fig.~\ref{fig:model} (right).
This class of latent structures is a popular choice for modeling a variety of problems such as human action recognition~\cite{yamato:92}, bitext word alignment~\cite{brown:93,vogel:96,blunsom:06}, POS tagging~\cite{merialdo:94,johnson:07}, acoustic modeling~\cite{jelinek:1997}, gene finding~\cite{lukashin:98}, and transliteration~\cite{reddy:09}, among others.  Importantly, we make no assumptions about conditional independence between any $\hid_i$ and $\mathbf{x}$.

Eq.~\ref{eq_parametric_form_pos} gives the parameteric form of the model for sequence labeling problems. $\boldsymbol{\lambda}$ and $\boldsymbol{\theta}$ are the parameters of the encoding and reconstruction models, respectively. $\mathbf{g}$ is a vector of clique-local feature functions.\footnote{We define $\hid_0$ to be a fixed ``start'' tag.}
\begin{align}
p_{\boldsymbol{\lambda},\boldsymbol{\theta}}(\mathbf{\rec}\mid\mathbf{\obs}) &= \sum_{\mathbf{\hid}} p_{\boldsymbol{\lambda}}(\mathbf{\hid}\mid\mathbf{\obs}) p_{\boldsymbol{\theta}}(\mathbf{\rec}\mid\mathbf{\hid}) =\sum_{\mathbf{\hid}  \in {\hiddomain}^{|\mathbf{\obs}|}} \frac{ e^{\sum_{i=1}^{|\mathbf{\obs}|} \boldsymbol{\lambda}^\top \mathbf{g}(\mathbf{\obs}, \hid_i, \hid_{i-1}, i)}}{ \sum_{\mathbf{\hid'}  \in {\hiddomain}^{|\mathbf{\obs}|}} e^{\sum_{i=1}^{|\mathbf{\obs}|} \boldsymbol{\lambda}^\top \mathbf{g}(\mathbf{\obs}, \hid'_i, \hid'_{i-1}, i)}} 
\prod_{i=1}^{|\mathbf{\obs}|} p_{\boldsymbol{\theta}}(\rec_i\mid \hid_i) \nonumber \\
&=\frac{ \sum_{\mathbf{\hid} \in {\hiddomain}^{|\mathbf{\obs}|}} \exp \left(\sum_{i=1}^{|\mathbf{\obs}|} \log \theta_{\rec_i \mid \hid_i} + \boldsymbol{\lambda}^\top \mathbf{g}(\mathbf{\obs}, \hid_i, \hid_{i-1}, i)\right)}{ \sum_{\mathbf{\hid'} \in {\hiddomain}^{|\mathbf{\obs}|}} \exp \sum_{i=1}^{|\mathbf{\obs}|} \boldsymbol{\lambda}^\top \mathbf{g}(\mathbf{\obs}, \hid'_i, \hid'_{i-1}, i)} \label{eq_parametric_form_pos}
\end{align}

\paragraph{Encoding and reconstruction.}
We model the \textit{encoding} part with a CRF, which allows us to exploit features with global scope in the structured observation $\mathbf{\obs}$, while keeping exact inference tractable (since the model does not generate $\mathbf{\obs}$, only conditions on it). The \textit{reconstruction} part, on the other hand,  grounds the model by generating a copy of the structured observations. We use a simple categorical (i.e., multinomial) distribution to independently generate $\rec_i$  given $\hid_i$.\footnote{Other possible parameterizations of the reconstruction model we would like to experiment with include a multivariate Gaussian for generating word embeddings and a na\"{\i}ve Bayes model for generating individual features of a word independently conditioned on the corresponding label. }
Fig.~\ref{fig:model} (right) is an instance of the model for POS induction with a sequential latent structure; each $\rec_i$ is generated from $p_{\boldsymbol{\theta}}(\rec_i \mid \hid_i)$.

We emphasize the importance of allowing modelers to define intuitive feature templates in a flexible manner. 
The need to efficiently add inductive bias via feature engineering has been the primary drive for developing CRF autoencoders.
For example, morphology, word spelling information, and other linguistic knowledge encoded as features were shown to improve POS induction \cite{smith:05}, word alignment \cite{dyer:11}, and other unsupervised learning problems.
The proposed model enables the modeler to define such features at a lower computational cost \textit{and} enables more expressive features with global scope in the structured input.
For example, we found that using predictions of other models as features is an effective method for model combination in unsupervised word alignment tasks, and found that conjoining sub-word-level features of consecutive words helps disambiguate their POS labels (see Appendix \ref{sec:appendix} for experimental details).

\paragraph{Extension: side information.}
Our model can be easily extended to condition on more context in the encoding part, the reconstruction part, or in both parts. Let $\boldsymbol{\sid}$ represent \textit{side information}: additional context which we may condition on in the encoding part, the reconstruction part, or both.  In our running example, side information could represent the set of possible tags for each word, a common form of weak ``supervision'' shown to help unsupervised POS learners.  In word alignment, where
$\hid_i = j$ indicates that $\obs_i$ translates to the $j$th source token, we treat the source sentence as side information, making its word forms available for feature extraction.

\paragraph{Extension: partial reconstruction.}
In our running POS example, the reconstruction model  $p_{\boldsymbol{\theta}}(\rec_i \mid \hid_i)$ defines a distribution over words given tags.  Because word distributions are heavy-tailed, estimating such a distribution reliably is quite challenging.  Our solution is to define a function $\pi : \obsdomain \rightarrow \recdomain$ such that $|\recdomain| \ll |\obsdomain|$, and let $\rec_i = \pi(\obs_i)$ be a deterministic transformation of the original structured observation. We can add indirect supervision by defining $\pi$ such that it represents observed information relevant to the latent structure of interest. For example, we found reconstructing Brown clusters \cite{brown:92} of tokens instead of their surface forms to improve POS induction. Other possible reconstructions include word embeddings, morphological and spelling features of words.

\paragraph{More general graphs.}  We presented the CRF autoencoder in terms of sequential Markovian assumptions for ease of exposition; however, this framework can be used to model arbitrary hidden structures. For example, instantiations of this model can be used for unsupervised learning of  parse trees~\cite{klein:04}, semantic role labels~\cite{swier:04}, and coreference resolution~\cite{poon:08} (in NLP), motif structures \cite{bailey:95} in computational biology, and object recognition \cite{weber:00} in computer vision.  The requirements for applying the CRF autoencoder model are:
\begin{itemizesquish}
\item An encoding discriminative model defining $p_{\boldsymbol{\lambda}}(\mathbf{\hid} \mid \mathbf{\obs}, \boldsymbol{\sid})$. The encoder may be any model family where \emph{supervised} learning from $\langle \mathbf{\obs}, \mathbf{\hid}\rangle$ pairs is efficient. 
\item A reconstruction model that defines $p_{\boldsymbol{\theta}}(\mathbf{\rec} \mid \mathbf{\hid}, \boldsymbol{\sid})$
such that inference over $\mathbf{\hid}$ given $\langle \mathbf{\obs}$, $\mathbf{\rec} \rangle$ is efficient.
\item The independencies among $\mathbf{\hid} \mid  \mathbf{\obs},\mathbf{\rec}$ are not strictly weaker than those among $\mathbf{\hid} \mid \mathbf{\obs}$. 
\end{itemizesquish}

\subsection{Learning \& Inference}
\label{sec:learning}
Model parameters are selected to maximize the regularized conditional log likelihood of reconstructed observations $\mathbf{\rec}$ given the structured observation $\mathbf{\obs}$:
\begin{equation}
\textstyle \ell\ell(\boldsymbol{\lambda}, \boldsymbol{\theta})  = R_1(\boldsymbol{\lambda}) + R_2(\boldsymbol{\theta}) + \sum_{(\mathbf{\obs,\rec}) \in {\cal{T}}}  \log \sum_{\mathbf{\hid}} p_{\boldsymbol{\lambda}}(\mathbf{\hid}\mid\mathbf{\obs}) \times p_{\boldsymbol{\theta}}(\mathbf{\rec}\mid\mathbf{\hid})
\end{equation}

We apply block coordinate descent, alternating between maximizing with respect to the CRF parameters ($\boldsymbol{\lambda}$-step) and the reconstruction parameters ($\boldsymbol{\theta}$-step).  Each $\boldsymbol{\lambda}$-step applies one or two iterations of a gradient-based convex optimizer.\footnote{We experimented with AdaGrad \cite{duchi:11} and L-BFGS. When using AdaGrad, we accummulate the gradient vectors across block coordinate ascent iterations. } 
The $\boldsymbol{\theta}$-step applies one or two iterations of EM~\cite{dempster:77}, with a closed-form solution in the M-step in each EM iteration.
The independence assumptions among $\mathbf{\hid}$ make the marginal inference required in both steps straightforward; we omit details for space.

In the experiments below, we apply a squared $L_2$ regularizer for the CRF parameters $\boldsymbol{\lambda}$, and 
a symmetric Dirichlet prior for categorical parameters $\boldsymbol{\theta}$. 

The asymptotic runtime complexity of each block coordinate descent iteration, assuming the first-order Markov dependencies in Fig.~\ref{fig:model} (right), is: 
\begin{align}
  \BigO{|\boldsymbol{\theta}| 
  + |\boldsymbol{\lambda}|  
  + |{\cal{T}}| \times |\mathbf{\obs}|_{\mathit{max}} \times |{\hiddomain}|_{\mathit{max}} 
    \times ( |{\hiddomain}|_{\mathit{max}} \times |F_{\hid_{i-1},\hid_i}| 
      +      |F_{\mathbf{\obs},\hid_i}| ) } \label{eq:runtime}
\end{align}
where $F_{\hid_{i-1},\hid_i}$ are the active ``label bigram'' features used in $\langle\hid_{i-1},\hid_i\rangle$ factors, 
$F_{\mathbf{\obs},\hid_i}$ are the active emission-like features used in $\langle \mathbf{\obs},\hid_i\rangle$ factors.
$|\mathbf{\obs}|_{\mathit{max}}$ is the maximum length of an observation sequence. 
$|{\hiddomain}|_{\mathit{max}}$ is the maximum cardinality\footnote{In POS induction, $|{\hiddomain}|$ is a constant, the number of syntactic classes which we configure to $12$ in our experiments. In word alignment, $|{\hiddomain}|$ is the size of the source sentence plus one, therefore $|{\hiddomain}|_{\mathit{max}}$ is the maximum length of a source sentence in the bitext corpus.} of the set of possible assignments of $\hid_i$.

After learning the $\boldsymbol{\lambda}$ and $\boldsymbol{\theta}$ parameters of the CRF autoencoder, test-time predictions are made using maximum a posteriori estimation, conditioning on both observations and reconstructions, i.e., 
$\hat{\mathbf{\hid}}_{\textrm{MAP}} = \arg \max_{\mathbf{\hid}} p_{\boldsymbol{\lambda},\boldsymbol{\theta}} (\mathbf{\hid} \mid \mathbf{\obs}, \mathbf{\rec})$. 

\section{Connections To Previous Work}
\label{sec:related} 

This work relates to several strands of work in unsupervised learning. Two broad types of models have been explored that support unsupervised learning with flexible feature representations.  Both are fully generative models that define joint distributions over $\mathbf{\obs}$ and $\mathbf{\hid}$.  We discuss these ``undirected'' and ``directed'' alternatives next, then turn to less closely related methods.

\subsection{Existing Alternatives for Unsupervised  Learning with Features}
\label{sec:alternatives}

\paragraph{Undirected models.} A Markov random field (MRF) encodes the joint distribution through local potential functions parameterized using features. Such models ``normalize globally,'' requiring during training the calculation of a partition function summing over all possible inputs and outputs. In our notation:
\begin{equation}
Z(\boldsymbol{\theta}) = \sum_{\mathbf{\obs} \in \obsdomain^\ast} \sum_{\mathbf{\hid} \in \hiddomain^{|\mathbf{\obs}|}} \exp \boldsymbol{\lambda}^\top \mathbf{\bar{g}}(\mathbf{\obs}, \mathbf{\hid}) \label{partition}
\end{equation}
where $\mathbf{\bar{g}}$ collects all the local factorization by cliques of the graph, for clarity.  The key difficulty is in the summation over all possible observations.  Approximations have been proposed, including contrastive estimation, which sums over subsets of $\obsdomain^\ast$ \cite{smith:05,vickrey:10} (applied variously to POS learning by \citet{haghighi:06} and word alignment by \citet{dyer:11}) and noise contrastive estimation \cite{mnih:12}.

\paragraph{Directed models.} The directed alternative avoids the global partition function by factorizing the joint distribution in terms of locally normalized conditional probabilities, which are parameterized in terms of features. For unsupervised sequence labeling, the model was called a ``feature HMM'' by \citet{berg-kirkpatrick:10}.
The local emission probabilities $p(\obs_i \mid \hid_{i})$ in a first-order HMM for POS tagging are reparameterized as follows (again, using notation close to ours):
\begin{equation}
p_{\boldsymbol{\lambda}}(\obs_i \mid \hid_i) = \frac{\exp \boldsymbol{\lambda}^\top \mathbf{g}(\obs_i, \hid_i)} {\sum_{\obs \in \obsdomain} \exp \boldsymbol{\lambda}^\top \mathbf{g}(\obs, \hid_i)}  \label{localpartition}
\end{equation}
The features relating hidden to observed variables must be local within the factors implied by the directed graph. We show below that this locality restriction excludes features that are useful (\S\ref{sec:pos}).

Put in these terms, the proposed autoencoding model is a hybrid directed-undirected model. 

\paragraph{Asymptotic Runtime Complexity of Inference.}
The  models just described cannot condition on arbitrary amounts of $\mathbf{\obs}$ without increasing inference costs.  Despite the strong independence assumptions of those models,
the computational complexity of inference required for learning with CRF autoencoders is better (\S\ref{sec:learning}).

Consider learning the parameters of an undirected model by maximizing likelihood of the observed data. Computing the gradient for a training instance $\mathbf{x}$ requires time
\begin{align}
  \BigO{
    |\boldsymbol{\lambda}|  
  + |{\cal{T}}| \times |\mathbf{\obs}| \times |{\hiddomain}| 
    \times ( |{\hiddomain}| \times |F_{\hid_{i-1},\hid_i}| 
    \textcolor{red}{ + |{\obsdomain}| \times |F_{\obs_i,\hid_i}|} ) }, \nonumber
\end{align}
where $F_{\obs_i-\hid_i}$ are the emission-like features used in an arbitrary assignment of $\obs_i$ and $\hid_i$. 
When the multiplicative factor $|{\obsdomain}|$ is large, inference is slow compared to CRF autoencoders. 

Inference in \emph{directed models} is faster than in undirected models, but still slower than CRF autoencoder models. 
In directed models \cite{berg-kirkpatrick:10}, each iteration requires time
\begin{align}
\BigO{
    |\boldsymbol{\lambda}|  
  + |{\cal{T}}| \times |\mathbf{\obs}| \times |{\hiddomain}| 
    \times ( |{\hiddomain}| \times |F_{\hid_{i-1},\hid_i}| 
      +      |F_{\obs_i,\hid_i}| )  
  \textcolor{red}{+ |\boldsymbol{\theta}'| \times \max(|F_{\hid_{i-1},\hid_i}|, |F_{{\obsdomain},\hid_i}|) }}, \nonumber  
\end{align}
where $F_{\obs_i,\hid_i}$ are the active emission features used in an arbitrary assignment of $\obs_i$ and $\hid_i$,
$F_{{\obsdomain}, {\hid}_i}$ is the union of all emission features used with an arbitrary assignment of $\hid_i$,
and $\boldsymbol{\theta}'$ are the local emission and transition probabilities. 
When $|\obsdomain|$ is large, the last term $|\boldsymbol{\theta}'| \times \max(|F_{\hid_{i-1},\hid_i}|, |F_{{\obsdomain},\hid_i}|)$ can be prohibitively large.

\subsection{Other Related Work}

The proposed CRF autoencoder is more distantly related  to several important ideas in less-than-supervised learning.

\paragraph{Autoencoders and other ``predict self'' methods.}
Our framework borrows its general structure, Fig.~\ref{fig:model} (left), as well as its name, from \emph{neural network} autoencoders. 
The goal of neural autoencoders has been to learn feature representations that improve generalization in otherwise supervised learning problems~\cite{vincent:2008,collobert:08,socher:10}. 
In contrast, the goal of CRF autoencoders is to learn specific \emph{interpretable} regularities of interest.\footnote{This is possible in CRF autoencoders due to the interdependencies among variables in the hidden structure and the manually specified feature templates which capture the relationship between observations and their hidden structures.}
It is not clear how {neural autoencoders} could be used to learn the latent structures that {CRF autoencoders} learn, without providing supervised training examples. \citet{stoyanov:11} presented a related approach for discriminative graphical model learning, including features and latent variables, based on backpropagation, which could be used to instantiate the CRF autoencoder.

\citet{daume:09b} introduced a reduction of an unsupervised problem instance to a series of single-variable supervised classifications. The first series of these construct a latent structure $\mathbf{\hid}$ given the entire $\mathbf{\obs}$, then the second series reconstruct the input.  The approach can make use of any supervised learner; if feature-based probabilistic models were used, a $|\obsdomain|$ summation (akin to Eq.~\ref{localpartition}) would be required.  On unsupervised POS induction, this approach performed on par with the undirected model of \citet{smith:05}.

\citet{minka:05} proposed cascading a generative model and a discriminative model, where {class labels} (to be predicted at test time) are {marginalized out} in the generative part first, and then (re)generated in the discriminative part. In CRF autoencoders, {observations} (available at test time) are {conditioned on} in the discriminative part first, and then (re)generated in the generative part.

\paragraph{Posterior regularization.}  Introduced by \citet{ganchev:10}, posterior regularization is an effective method for specifying constraint on the posterior distributions of the latent variables of interest; a similar idea was proposed independently by \citet{bellare:09}.  For example, in POS induction, every sentence might be expected to contain at least one verb.  This is imposed as a soft constraint, i.e., a feature whose expected value under the model's posterior  is constrained. Such expectation constraints are specified directly by the domain-aware model designer.\footnote{In a semi-supervised setting, when some labeled examples of the hidden structure are available, \citet{druck:10} used labeled examples to estimate desirable expected values.  We leave semi-supervised applications of CRF autoencoders to future work; see also \citet{suzuki:08}.}
The approach was applied to unsupervised POS induction, word alignment, and parsing.
Although posterior regularization was applied to directed feature-less generative models, the idea is orthogonal to the model family and can be used to add more inductive bias for training CRF autoencoder models.

\begin{figure}
\centering   {\includegraphics[clip=true,trim = 0in .6in 0in 0.6in, height=2.5in]{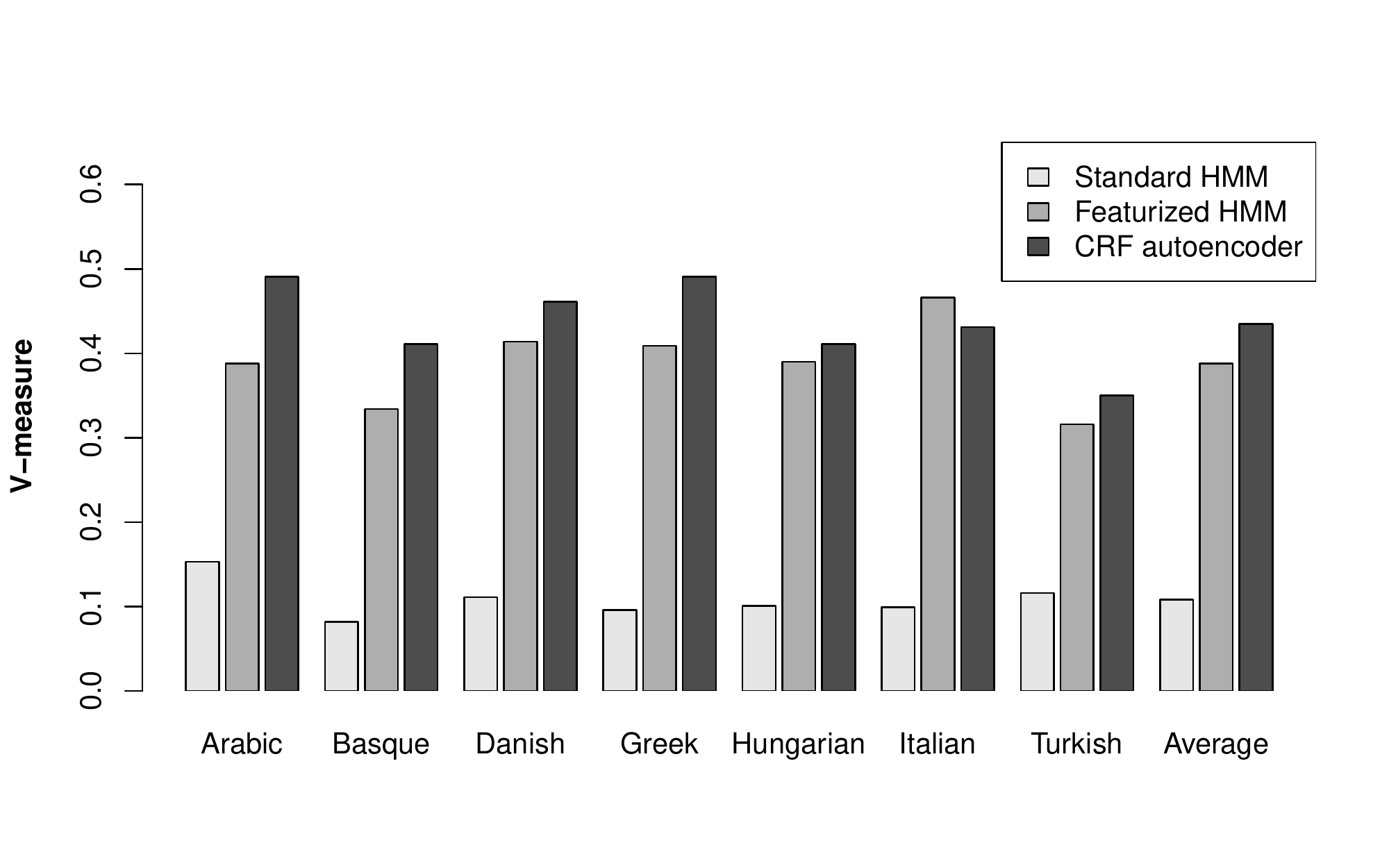}}
  \caption{V-measure \citep{rosenberg:07} of induced parts of speech in seven languages. The CRF autoencoder with features spanning multiple words and with Brown cluster reconstructions achieves the best results in all languages but Italian, closely followed by the feature-rich HMM of \citet{berg-kirkpatrick:10}. The standard multinomial HMM consistently ranks last.\label{fig:vmeasure}}
\end{figure}

\section{Evaluation}
\label{sec:exp}
We evaluate the effectiveness of CRF autoencoders for learning from unlabeled examples in POS induction and word alignment.  
We defer the detailed experimental setup to Appendix \ref{sec:appendix}.

\paragraph{Part-of-Speech Induction Results.}
Fig.~\ref{fig:vmeasure} compares predictions of the CRF autoencoder model in seven languages 
to those of a featurized first-order HMM model \citep{berg-kirkpatrick:10} and a standard (feature-less) first-order HMM, 
using  V-measure  \citep{rosenberg:07} (higher is better). 
First, note the large gap between both feature-rich models on the one hand, and the feature-less HMM model on the other hand.
Second, note that CRF autoencoders outperform featurized HMMs in all languages, except Italian, with an average relative improvement of 12\%.

These results provide empirical evidence that feature engineering is an important source of inductive bias for unsupervised structured prediction problems.
In particular, we found that using Brown cluster reconstructions  and specifying features which span multiple words significantly improve the performance. 
Refer to Appendix A for more analysis.

\paragraph{Bitext Word Alignment Results.}
First, we consider an intrinsic evaluation on a Czech-English dataset of manual alignments, measuring the alignment error rate (AER; \cite{och:03}).  
We also perform an extrinsic evaluation of translation quality in three language pairs, using case-insensitive Bleu~\cite{papineni:02} 
of a machine translation system ({cdec}\footnote{\url{http://www.cdec-decoder.org/}}~\cite{dyer:10}) built using the word alignment predictions of each model.   

AER for variants of each model (forward, reverse, and symmetrized) are shown in Table~\ref{table:aer} (left).  Our model significantly outperforms both baselines.
Bleu scores on the three language pairs are shown in Table~\ref{table:bleu}; alignments obtained with our CRF autoencoder model improve translation quality of the Czech-English and Urdu-English translation systems, but not of Chinese-English.  This is unsurprising, given that Chinese orthography does not use letters, so that source-language spelling and morphology features our model incorporates introduce only noise here.  Better feature engineering, or more data, is called for. 

We have argued that the feature-rich CRF autoencoder will scale better than its feature-rich alternatives.  
Fig.~\ref{fig:scalability} (in Appendix \ref{sec:wa}) shows the average per-sentence inference runtime for the CRF autoencoder compared to exact inference in an MRF \cite{dyer:11} with a similar feature set, as a function of the number of sentences in the corpus. 
For CRF autoencoders, the average inference runtime  grows slightly due to the increased number of parameters, while it grows substantially with vocabulary size in MRF models \cite{dyer:11}.\footnote{We only compare runtime, instead of alignment quality, because retraining the MRF model with exact inference was too expensive.}

\begin{table}
  \begin{minipage}{.5\textwidth}
    \centering
    \begin{tabular}{r|ccc}
      \hline
      direction   &  \fastalign   & \modelfour & \auto  \\
      \hline
      forward     & 27.7         &   31.5    & \textbf{27.5}    \\
      reverse    & 25.9         &   24.1    & \textbf{21.1}    \\
      symmetric   & 25.2         &   22.2    & \textbf{19.5}       \\
      \hline
    \end{tabular}
  \end{minipage}
  \begin{minipage}{.5\textwidth}
    \centering
    \begin{tabular}{r|ccc}
      \hline
      pair &\fastalign&\modelfour&\auto\\
      \hline
      cs-en   &15.2{\tiny${\pm0.3}$}&15.3{\tiny$\pm0.1$}&\textbf{15.5}{\tiny$\pm0.1$}\\
      ur-en    &20.0{\tiny$\pm0.6$}&20.1{\tiny$\pm0.6$}&\textbf{20.8}{\tiny$\pm0.5$}\\
      zh-en &\textbf{56.9}{\tiny$\pm1.6$}&56.7{\tiny$\pm1.6$}&56.1{\tiny$\pm1.7$}\\
      \hline
    \end{tabular}
  \end{minipage}
  \caption {Left: AER results (\%) for Czech-English word alignment. Lower values are better. \label {table:aer}. Right: Bleu translation quality scores (\%) for Czech-English, Urdu-English and Chinese-English. Higher values are better. \label {table:bleu}. }  
\end{table}

\section{Conclusion}
\label{sec:conclusion}
We have presented a general and scalable framework to learn from unlabeled examples for structured prediction. The technique allows
features with global scope in observed variables with favorable asymptotic inference runtime.  We achieve this by embedding a CRF as the encoding model in the input layer of an autoencoder, and reconstructing a transformation of the input at the output layer using simple categorical distributions. The key advantages of the proposed model are scalability and modeling flexibility. We applied the model to POS induction and bitext word alignment, obtaining results that are competitive with the state of the art on both tasks.

\subsubsection*{Acknowledgments}
We thank Brendan O'Connor,  Dani Yogatama, Jeffrey Flanigan,  Manaal Faruqui, Nathan Schneider, Phil Blunsom and the anonymous 
reviewers for helpful suggestions.
We also thank Taylor Berg-Kirkpatrick for providing his implementation of the POS induction baseline, 
and Phil Blunsom for sharing POS induction evaluation scripts.
This work was sponsored by the U.S.~Army Research Laboratory and the U.S.~Army Research Office
under contract/grant number W911NF-10-1-0533.
The statements made herein are solely the responsibility of the authors.

{
\bibliographystyle{abbrvnat}
\DeclareRobustCommand{\VAN}[3]{#3}
\setlength{\bibsep}{1mm}
\bibliography{biblio}
}

\newpage

\appendix

\section{Experimental Details}
\label{sec:appendix}
In this section, we describe our experimental setup for part of speech induction and bitext word alignment in some detail.

\subsection{Part-of-Speech Induction Experimental Setup}
\label{sec:pos}
The first task, part-of-speech (POS) induction, is a classic NLP problem which aims at discovering syntactic classes of tokens in a monolingual corpus, with a predefined number of classes. 
An example of a POS-tagged English sentence is in Fig.~\ref{fig:examples}.

\paragraph{Data.}
We use the plain text from CoNLL-X~\cite{buchholz:06} and CoNLL 2007~\cite{nivre:07} training data in seven languages to train the models: Arabic, Basque, Danish, Greek, Hungarian, Italian and Turkish. 
For evaluation, we obtain gold-standard POS tags by deterministically mapping the language-specific POS tags from the shared task training data to the corresponding universal POS tag set
\cite{petrov:12}.\footnote{\url{http://code.google.com/p/universal-pos-tags/}}

\paragraph{Setup.} 
We configure our model (as well as baseline models) to induce $|{\hiddomain}|=12$ classes. 
We use zero initialization of the CRF parameters, and initialize the reconstruction model parameters with a basic first-order HMM.
In each block-coordinate ascent iteration, we run one L-BFGS iteration (including a line search) to optimize $\boldsymbol{\lambda}$, followed by one EM iteration to optimize $\boldsymbol{\theta}$. 
Training converges after $70$ block-coordinate ascent iterations. 

\paragraph{Evaluation.}
Since we do not use a tagging dictionary, the word classes induced by our model are not identifiable. 
We use two cluster evaluation metrics commonly used for POS induction:
(i) many-to-one \cite{johnson:07}, which infers a mapping across the syntactic clusters in the gold vs.~predicted labels (higher is better); and
(ii) V-measure \cite{rosenberg:07}, an entropy-based metric which explicitly measures the homogeneity and completeness of predicted clusters (again, higher is better). 

\paragraph{CRF Autoencoder Model Instantiation.}
Table \ref{table:spec} (right) describes the symbols and variables we use in context of the POS induction problem.
We use a first-order linear CRF for the encoding part with the following feature templates:
\begin{itemizesquish}
\item $\langle y_i, y_{i-1} \rangle, \forall i$
\item $\langle y_i, \text{sub}_j(x_i) \rangle, \forall i, j$
\item $\langle y_i, \text{sub}_j(x_i), \text{sub}_k(x_{i-1}) \rangle, \forall i, j, k$
\item $\langle y_i, \text{sub}_j(x_i), \text{sub}_k(x_{i+1}) \rangle, \forall i, j, k$
\end{itemizesquish}

Where $\text{sub}_j(x_i)$ is one of the following sub-word-level feature percepts:
\begin{itemizesquish}
\item Prefixes and suffixes of lengths two and three, iff the affix appears in more than 0.02\% of all word types,
\item Whether the word contains a digit,
\item Whether the word contains a hyphen,
\item Whether the word starts with a capital letter,
\item Word shape features which map sequences of the same character classes into a single character (e.g., `McDonalds' $\rightarrow$ `AaAa', `-0.5' $\rightarrow$ `\#0\#0'),
\item The lowercased word, iff it appears more than 100 times in the corpus.
\end{itemizesquish}

In the reconstruction model, we generate the Brown cluster of a word \cite{brown:92},\footnote{We use \thin{brown-cluster -c 100} v1.3, available at \texttt{https://github.com/percyliang/brown-cluster} \cite{liang:05} with data from \url{http://corpora.informatik.uni-leipzig.de/}.} 
conditioned on the POS tag using a categorical (i.e., multinomial) distribution.
No side information is used in this model instantiation.
Predictions are the best value of the latent structure according to the posterior $p(\mathbf{\hid} \mid  \mathbf{\obs}, \mathbf{\rec}, \boldsymbol{\sid})$.

\paragraph{Baselines.}
We consider two baselines:
\begin{itemizesquish}
\item \hmm:  a standard first-order hidden Markov model learned with EM;\footnote{Among 32 Gaussian initializations of model parameters, we use the HMM model which gives the highest likelihood after 30 EM iterations. }
\item \fhmm:  the directed alternative discussed in the main paper, as implemented by \citet{berg-kirkpatrick:10}, with the feature set from \citet{smith:05}.
\end{itemizesquish}

\paragraph{Tuning Hyperparameters.}
In our models, we use a squared $L_2$ regularizer for CRF parameters $\boldsymbol{\lambda}$, and a symmetric Dirichlet prior for categorical parameters $\boldsymbol{\theta}$ with the same regularization strength for all languages.
The \fhmm{} baseline also uses a squared $L_2$ regularizer for the log-linear parameters.
The hyperparameters of our model, as well as baseline models, were tuned to maximize many-to-one accuracy for The English PennTreebank. 
The \fhmm{} model uses $L_2$ strength $=0.3$.
The \auto{} model uses $L_2$ strength $=2.5$, $\alpha=0.1$. 

\begin{figure}
  \centering
  \includegraphics[clip=true,trim=0in 0.6in 0in 0.6in,height=2.5in]{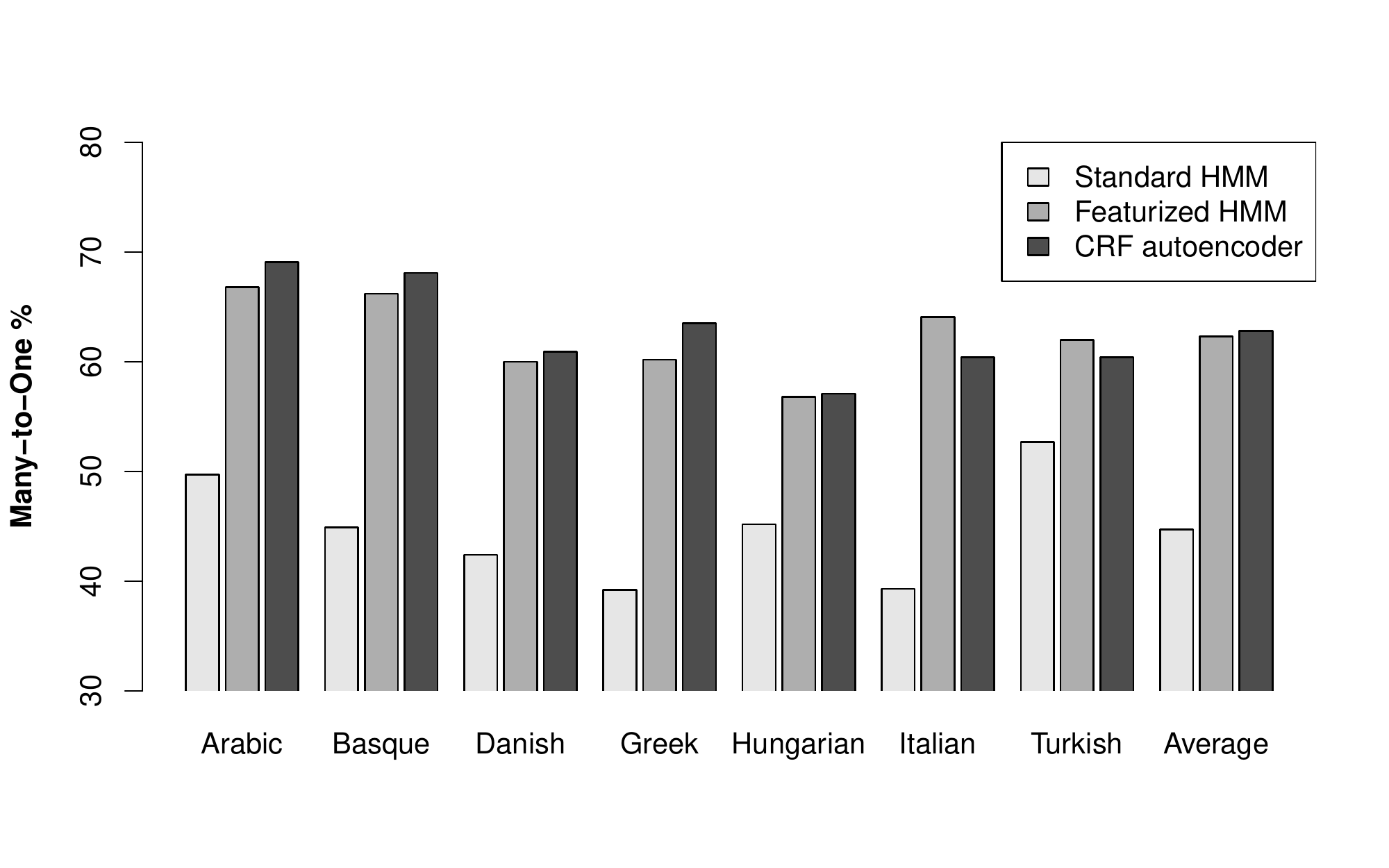}\\
  \caption{Many-to-one accuracy  of POS induction. \label{fig:manytoone}}
\end{figure}

\subsection{Bitext Word Alignment Experimental Setup}
\label{sec:wa}
Word alignment is an essential step in the training pipeline of most statistical machine translation systems~\cite{koehn:2010}. Given a sentence in the source language and its translation in the target language, the task is to find which {source} token, if any, corresponds to each token in the {target} translation. We make the popular assumption that each token in the target sentence corresponds to zero or one tokens in the source sentence. Fig.~\ref{fig:examples} shows a Spanish sentence and its English translation with word alignments.
As shown in Table~\ref{table:spec} (right), an observation $\mathbf{\obs}$ consists of tokens in the target sentence, while side information $\boldsymbol{\sid}$ are tokens in the source sentence.
Conditioned on a source word, we use a categorical  distribution to generate the corresponding target word according to the inferred alignments.

\paragraph{Data.}
We consider three language-pairs:   Czech-English, Urdu-English, and Chinese-English. 
For Czech-English, we use 4.3M bitext tokens for training from the NewsCommentary corpus, WMT10 data set for development, and WMT11 for testing. For Urdu-English, we use the train (2.4M bitext tokens), development, and test sets provided for NIST open MT evaluations 2009. For Chinese-English, we use the BTEC train (0.7M bitext tokens), development, and test sets (travel domain).

\begin{SCfigure}[][t]
  \caption{Average inference runtime per sentence pair for word alignment in seconds ($y$-axis), as a function of the number of {sentences} used for training ($x$-axis). \label{fig:scalability}}
  \includegraphics[height=2.5in]{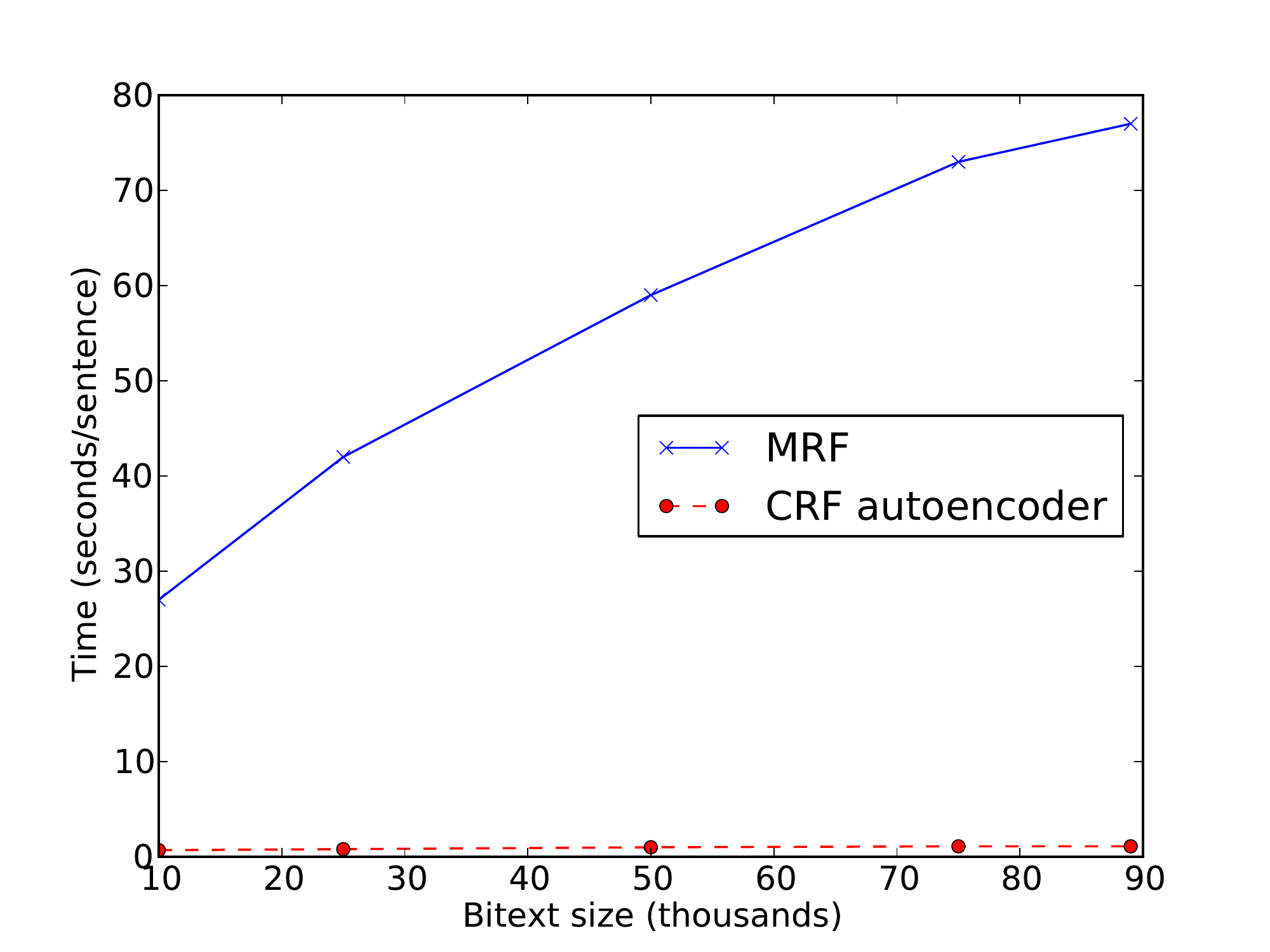}
\end{SCfigure}

\paragraph{CRF Autoencoder Model Instantiation.}
For word alignment, we define the reconstruction model as follows: $p_{\boldsymbol{\theta}}(\mathbf{\rec}\mid\mathbf{\hid},\boldsymbol{\sid}) = \prod_{i=1}^{|\mathbf{\obs}|} \theta_{\rec_i|\sid_{\hid_i}}$, where $\rec_i$ is the Brown cluster\footnote{We again use~\cite{liang:05} with $80$ word classes.} of the word at position $i$ in the target sentence. We use a squared $L_2$ regularizer for the log-linear parameters $\boldsymbol{\lambda}$ and a symmetric Dirichlet prior for the categorical parameters $\boldsymbol{\boldsymbol{\theta}}$ with the same regularization strength for all language pairs ($L_2$ strength $= 0.01$, Dirichlet $\alpha=1.5$). The hyperparameters were optimized to minimize alignment error rate ({AER}) on a development dataset of French-English bitext. The reconstruction model parameters $\boldsymbol{\theta}$ are initialized with the parameters taken from IBM Model 1 after five EM iterations~\cite{brown:93}. In each block-coordinate ascent iteration, we use L-BFGS to optimize $\boldsymbol{\lambda}$, followed by two EM iterations to optimize $\boldsymbol{\theta}$. Training converges when the relative improvement in objective value falls below $0.03$ in one block-coordinate ascent iteration, typically in less than $10$ iterations of block-coordinate ascent.

We follow the common practice of training two word alignment models for each dataset, one with English as the target language (forward) and another with English as the source language (reverse). We then use the \thin{grow-diag-final-and} heuristic~\cite{koehn:03} to symmetrize alignments before extracting translation rules. 

\paragraph{Features.}
We use the following features: deviation from diagonal word alignment $|\frac{\hid_i}{|\boldsymbol{\sid}|} - \frac{i}{|\mathbf{\obs}|}|$; log alignment jump $\log |\hid_i-\hid_{i-1}|$; agreement with forward, reverse and symmetrized baseline alignments of \giza{} and \fastalign; Dice measure of the word pair $\obs_i$ and $\sid_{\hid_i}$;  difference in character length between $\obs_i$ and $\sid_{\hid_i}$; orthograhpic similarity between $\obs_i$ and $\sid_{\hid_i}$, punctuation token aligned to a non-punctuation token; punctuation token aligned to an identical token; 4-bit prefix of the Brown cluster of $\obs_i$ conjoined with 4-bit prefix of the Brown cluster of $\sid_{\hid_i}$; forward and reverse probability of the word pair $\obs_i, \sid_{\hid_i}$ with \fastalign, as well as their product. The outputs of other unsupervised aligners are standard (and important!)~features in supervised CRF aligners \cite{blunsom:06}; however, they are nonsensical in a joint model over alignments and sentence pairs.

\paragraph{Baselines.}
Due to the cost of estimating feature-rich generative models for unsupervised word alignment on the data sizes we are using (e.g., \fhmm{} and \thin{dyer-11}), we only report the per-sentence computational cost of inference on these baselines. For alignment quality baselines, we report on results from two state-of-the-art baselines that use multinomial parameterizations which support M-step analytic solutions, rather than feature-rich parameterizations:
\fastalign~\cite{dyer:13}\footnote{\texttt{https://github.com/clab/fastalign}} and \modelfour~\cite{brown:93}. \fastalign~is a recently proposed reparameterization of IBM Model 2~\cite{brown:93}. \modelfour, as implemented in \giza~\cite{gao:08} is the most commonly used word alignment tool in state-of-the-art machine translation systems. 

\paragraph{Evaluation.}
When gold standard word alignments are available (i.e., for Czech-English), we use {AER}~\cite{och:03} to evaluate the alignment predictions of each model. We also perform an extrinsic evaluation of translation quality for all data sets, using case-insensitive {Bleu}~\cite{papineni:02} 
of a hierarchical MT system built using the word alignment predictions of each model.

\end{document}